\pdfoutput=1
\documentclass[11pt]{article}

\usepackage[final]{acl}
\usepackage{times}
\usepackage{latexsym}

\usepackage[T1]{fontenc}
\usepackage[utf8]{inputenc}

\usepackage{microtype}

\usepackage{inconsolata}

\usepackage{graphicx}

\usepackage{booktabs}
\usepackage{adjustbox}
\usepackage{amssymb}
\usepackage{expex}
\usepackage{multirow}
\usepackage{multicol}

\title{BabyLM Challenge: Exploring the Effect of Variation Sets \\ on Language Model Training Efficiency}

\author{Akari Haga$^{1}$\quad Akiyo Fukatsu$^{2}$\quad Miyu Oba$^{1}$\quad Arianna Bisazza$^{3}$\quad Yohei Oseki$^{2}$ \\
 $^1$Nara Institute of Science and Technology\\
 $^2$The University of Tokyo\\
 $^3$University of Groningen\\
\texttt{\{haga.akari.ha0, oba.miyu.ol2\}@is.naist.jp} \\
\texttt{\{akiyofukatsu, oseki\}@g.ecc.u-tokyo.ac.jp} \\
\texttt{a.bisazza@rug.nl}
}

\begin{document}
\maketitle
\begin{abstract}
While current large language models have achieved a remarkable success, their data efficiency remains a challenge to overcome. 
Recently it has been suggested that child-directed speech (CDS) can improve training data efficiency of modern language models based on Transformer neural networks. However, it is not yet understood which specific properties of CDS are effective for training these models. 
In the context of the BabyLM Challenge, we focus on Variation Sets (VSs), sets of consecutive utterances expressing a similar intent with slightly different words and structures, which are ubiquitous in CDS.
To assess the impact of VSs on training data efficiency, 
we augment CDS data with different proportions of artificial VSs and use these datasets to train an auto-regressive model, GPT-2.
We find that the best proportion of VSs depends on the evaluation benchmark: BLiMP and GLUE scores benefit from the presence of VSs, but EWOK scores do not.
Additionally, the results vary depending on multiple factors such as the number of epochs and the order of utterance presentation.
Taken together, these findings suggest that VSs can have a beneficial influence on language models, while leaving room for further investigation.

\end{abstract}

\section{Introduction}
While current language models (LMs) demonstrate outstanding performance in a range of linguistic and reasoning tasks, there is ample scope to enhance their data efficiency.
A state-of-the-art LM like Chinchilla uses as much as 1.4 trillion words for pretraining, whereas humans master their native language by hearing less than 100M words by the age of 13~\citep{warstadt2022artificial}.

Child language acquisition could provide insights into it, given that children acquire basic grammar by the age of six~\citep{paul-1981, Kemp2005-xf}, without as varied and abundant linguistic inputs as those given to modern LMs. 
Various studies argue that this highly efficient learning is aided by children's limited cognitive abilities and specific types of inputs towards children~\citep{NEWPORT199011, fernald-1985, jusczyk-1997, rowe-2012, KEMPE2024101121}. 
Inspired by this, the BabyLM Challenge aims at improving data efficiency in language models as well as providing insights into child language acquisition. 

It is also suggested that CDS is a preferable domain for facilitating the acquisition of linguistic knowledge compared to other domains of data.
The findings of these studies include efficient pretraining without sacrificing the performance~\citep{huebner-etal-2021-babyberta}, enhanced semantic extraction~\citep{you-etal-2021}, and superior induction of hierarchical structure~\citep{mueller-linzen-2023-plant}.
While these studies suggest that CDS helps LMs learn from limited datasets, further research is needed to determine which specific properties of CDS provide an advantage to LMs.
 
As one of such properties, some studies highlight Variation Sets (VSs), which are sets of (mostly consecutive) utterances expressing a similar intent with slight variations in the use of words and structures~\citep{kuntay-slobin-1996}.
This specific pattern is ubiquitous in CDS, but not in other speech genres.
In first and second language acquisition, several studies indicate that VSs in CDS support learning of syntactic structure~\citep{hoff-ginsberg-1990,brodsky-waterfall-2007,onnis-2008} by maintaining children's attention on the circumscribed topic and promoting comprehension by introducing new information~\citep{lester-etal-2022}.
These findings suggest that VSs are beneficial for language learning in general and thus could enhance the learning process in LMs.

In this work, we explore this hypothesis by examining the effect of VSs on language models' data efficiency.
To fully control the impact of VSs, we construct artificial VSs based on the description by~\citet{kuntay-slobin-1996}, mixing it with actual CDS at various rates (0\%, 20\%, 40\%, 60\%, 80\%, 100\%).
Then we compare the models' accuracy on these constructed datasets and shuffled datasets on BLiMP~\citep{warstadt-etal-2020-blimp-benchmark}, EWOK~\citep{ivanova2024elementsworldknowledgeewok}, and GLUE~\citep{wang-2018-glue}. 

\section{Related Work}
\subsection{Child-directed Speech}
CDS is a specific speech genre that parents and other caregivers use to address children, and that differs from adult-directed speech (ADS). 
CDS usually has simpler sentence structures, more repetitive speech, and more limited vocabulary~\citep{snow-1972, farwell-1975, fernald-et-al-1989, kirchhoff-schimmel-2005}. 

Studies in child language development suggest that this specific speech genre is necessary for successful language acquisition among children.
For example, ~\citet{fernald-1985} tests 48 four-month-old infants on operant auditory preference procedure and finds that they preferred CDS to ADS.
~\citet{jusczyk-1997} reports that infants can segment speech better when they hear CDS than ADS.
~\citet{rowe-2012} conducts a longitudinal study on 50 parent--child dyads, demonstrating that parents' sophisticated vocabulary and decontextualized (narrative) conversation accelerate later vocabulary development in children.\footnote{Note that in several cultures, CDS is infrequent.~\citep{cristia2019child, weber2017cultural}.}

Following the BabyLM setup, we do not work with speech but with textual transcriptions of CDS. While sacrificing the richness of the speech signal, this choice makes the task accessible to a wider audience of computational linguistics researchers, by reducing the data complexity of the input. Henceforth, we will use CDS to denote textual transcriptions of child-directed speech.

\section{Computational Studies on CDS}
Computational studies further investigate whether CDS is beneficial for acquiring grammatical knowledge in models as well as for human language acquisition. 
\citet{huebner-etal-2021-babyberta} demonstrate that the use of child-directed speech (CDS) enables a small-sized RoBERTa~\citep{Liu2019RoBERTaAR} model trained on 5M words to attain similar linguistic competence as a RoBERTa trained on 30B words. 
~\citet{you-etal-2021} examine that CDS has rich semantic information for grasping causal semantics without syntactic structures, finding that CDS is effective in learning to extract semantic information.
Furthermore,~\citet{mueller-linzen-2023-plant} argue that LMs can induce hierarchical structures better when trained on CDS than on other typical datasets like Wikipedia.

While these findings demonstrate the positive effect of CDS on language learning, we are interested in which specific properties of CDS contribute to this effect.
One of the reasons why CDS can enhance LMs' acquisition of syntactic structures could be its lower lexical complexity, i.e., fewer word types ~\citep{mueller-linzen-2023-plant}, which stems from the high repetitiveness of items in CDS. However, this repetition occurs across multiple utterances, a characteristic unique to CDS.
We hypothesize that this could be a key factor in the success of LMs' language learning.

\subsection{Variation Sets}
In studies of first and second language acquisition, VSs~\citep{kuntay-slobin-1996} have gained attention as a key factor in successful language development. \citet{kuntay-slobin-1996} describe the characteristics of VSs as follows: in successive utterances, 1) the same content is repeated or rephrased, 2) the semantic intent remains consistent, and 3) operations such as word substitution, phrase addition or deletion, and phrase reordering occur. An example of a typical VS in English is provided by \citet[][p.44]{wiren-etal-2016-longitudinal}:

\begin{enumerate}
    \item[(1)]
    You can put the animals there. \\
    You can take the pig and the cat and put them there. \\
    Can you put them there? \\
    Good. \\
    Can you put the pig there too? 
\end{enumerate}

Several studies suggest that VSs indeed enhance language learning. For example,~\citet{hoff-ginsberg-1990} argue that repeating identical utterances boosts children's syntactic development, while consecutive utterances with slight variations provide clues about sentence structure, aiding syntactic development.
~\citet{brodsky-waterfall-2007} conduct a corpus-based study and demonstrate that utterances with partial repetitions, such as VSs, can be overly information-dense for learners.
~\citet{onnis-2008} investigate the effect of VSs in language learning by teaching adults an artificial language. Their results show that VSs help adult learners parse sentences, suggesting that comparing consecutive sentences provides clues for learning syntactic structures.
Taken together, these findings suggest that seeing contextually consistent utterances with slight differences in wording could make structural differences more salient, leading to better prediction for syntactic structure in LMs.

\section{Method}
\begin{figure*}[tb]
    \centering
    \includegraphics[width=\linewidth]{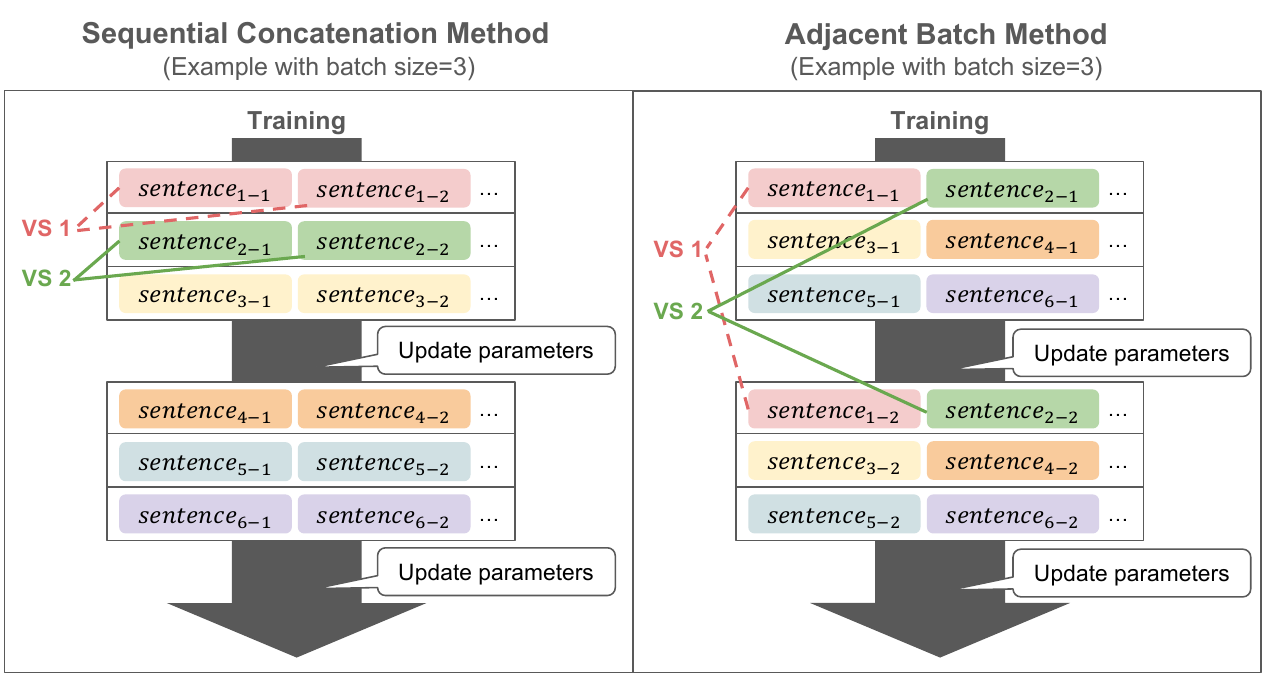}
    \caption{Two methods for inputting VSs to the model during training. Each figure illustrates an example with a batch size of 3. The figure on the left shows the method of concatenating VSs into a single sequence. In this setting, the model always processes the sentences within a VS sequentially. The figure on the right shows the method of distributing each sentence of a VS into adjacent batches. In this setting, the model updates its parameters after observing each sentence in the VS before proceeding to the next sentence in the same set. In the figures, sentence $i$-$j$ indicates the $j$-th sentence in the $i$-th VS.}
    \label{fig:methods}
\end{figure*}

Inspired by human studies of language acquisition, we want to examine whether VSs can also help a language model recognize sentence structures in the language. 
To our knowledge, this effect has only been explored in the pilot experiment of \citet{katano-2024}.
The experiment consisted of extracting naturally occurring VSs from CDS data using multiple automatic VS detection methods. The results showed no significant effect of VSs on syntactic performance measured on BLiMP,
but this could be due to the difficulty in fully controlling the number of actual VSs detected automatically~\citep{lester-etal-2022}.

To address this difficulty, we opt for the use of synthetic VSs, which allows us to fully control the proportion of VSs in our training dataset.
For this purpose, we use~\texttt{gpt4o-mini}\footnote{https://openai.com/index/gpt-4o-mini-advancing-cost-efficient-intelligence/} to generate artificial VSs and augment the training data in different proportions.\footnote{To fully control the proportion of VSs within the dataset, we shuffled all sentences except for the artificial VSs to ensure that no natural VSs are included.}
While the use of a large LM seems to contradict the goal of improving data efficiency, we see this as a first step to measuring the importance of VSs. In case of successful results, future work could explore less costly methods to generate VSs, such as template- or syntactic rule-based.

Humans hear sentences in a VS sequentially. However, it is not clear how to present VSs to a model in a way that is equivalent to human input, nor how to maximize the effect of VSs.
Therefore, we conduct experiments using two methods.
The first method, as shown in the example on the left side of Figure~\ref{fig:methods}, involves concatenating the VS into a single instance and providing it to the model. In this configuration, the model is forced to sequentially process the sentences within the VSs. We named this method the ``Sequential Concatenation Method''.
The second method, as shown in the example on the right side of Figure~\ref{fig:methods}, involves placing each sentence in the VS in adjacent batches. In this configuration, the model updates its parameters after processing one sentence in the VS before moving on to the next sentence within the same set. We named this method the ``Adjacent Batch Method''.

\subsection{Model Architecture}
It has been suggested that children use predictive sentence processing, actively integrating syntactic and semantic information to foresee the upcoming categories of words~\citep{Borovsky2012-oj}. Recent studies suggest that children's predictive behavior aids their language acquisition~\citep{Reuter2019-xu}. 
These findings suggest that predictive processing is a powerful tool for learning sentence structure.
Thus, we use GPT-2~\citep{radford2018improving}, an auto-regressive (left-to-right) language model rather than a bidirectional one like BERT~\citep{devlin-etal-2019-bert}. Hyperparameters are shown in Appendix~\ref{tab:hypara}. 

\subsection{Synthesizing Variation Sets}
\begin{table}[t]
    \begin{adjustbox}{width=\columnwidth,center}  
    \centering
    \begin{tabular}{ll}
        \toprule
            Original CDS & Generated VS \\
            \midrule
            What do you want?  &  \begin{tabular}[c]{l@{}l@{}l@{}l@{}l@{}}What do you need? \\ What do you want to have? \\ Can you tell me what you want? \\ What is it that you want? \\ What do you feel like getting? \end{tabular}\\
            \midrule
            \begin{tabular}[c]{l@{}l@{}}What did Laura do \\ last night? \end{tabular}&  \begin{tabular}[c]{l@{}l@{}l@{}l@{}l@{}} What did Laura do yesterday evening?  \\
            What was Laura doing last night? \\
            Can you tell me what Laura did last night? \\
            What activity did Laura have last night? \\
            What was Laura up to last night?  
 \end{tabular} \\
        \bottomrule
    \end{tabular}
    \end{adjustbox}
    \caption{Examples of VS generated by~\texttt{gpt4o-mini}. The left column presents the original sentence in CDS, and the right column presents artificial VSs generated by the model.}
    \label{tab:example}
\end{table}
To construct training data, we extract 10 million words of CDS from English corpora in CHILDES~\citep{macwhinney2000}. 
We eliminate utterances consisting of less than three words.
In the previous literature, VSs were extracted from CDS, although these VSs contain some intervening utterances within a set of VSs:

\begin{enumerate}
    \item[(2)]
    You wanna straw? \\
    Here's your straw. \\
    \textbf{Uh oh.} \\
    Where's the straw? \\
\end{enumerate}

Children can efficiently ignore these intervening utterances, whereas these utterances can be noisy for language models. 
Given that we intend to explore the impact of speech patterns described by~\citet{kuntay-slobin-1996} on language models, we develop artificial data to eliminate the potential noise.
For developing artificial VSs, we use \texttt{gpt4o-mini} and ask the model to generate a set of utterances that correspond to the descriptions by~\citet{kuntay-slobin-1996} and a prototypical example (see full prompt in Appendix~\ref{sec:appendix-prompt}). Table~\ref{tab:example} shows examples of original utterances in CDS and generated VSs based on them.
Approximately 48\% of the generated VSs are questions.

By using the artificial data, we can examine the upper bound of the influence of VSs on learning by language models.

\subsection{Composing the Datasets}
We mix artificial VSs with shuffled CDS since CDS includes a certain percentage of VS~\citep{waterfall-2006, brodsky-waterfall-2007, onnis-2008}. 
The percentage depends on corpora.
To explore at which ratio VSs should be mixed with CDS to enhance the model’s learning, we mix VSs with CDS at various ratios (0, 20, 40, 60, 80, 100).
We shuffled all CDS except for the artificial VSs to ensure that no natural VSs are included.
We then feed the model with these datasets using two different methods: concatenating each VS into a single sequence (``Sequential Concatenation Method'') and placing each sentence in VS in adjacent batches (``Adjacent Batch Method''). 

\subsection{Evaluation}
To disentangle the effect of the \textit{presence} of rephrasings (or variations) of the same sentence in the data from their consecutive \textit{order} of presentation, we compare the results of each VS dataset with the shuffled version of the same dataset.

We evaluate models on BLiMP~\citep{warstadt-etal-2020-blimp-benchmark} and its supplemental tasks, EWOK~\citep{ivanova2024elementsworldknowledgeewok}, and GLUE~\citep{wang-2018-glue} using the evaluation pipeline provided by the BabyLM organizers~\citep{babylm-2024, eval-harness}.
BLiMP and EWOK are used for zero-shot evaluation, whereas GLUE is used for fine-tuning evaluation.
BLiMP is a binary classification task for evaluating grammatical knowledge in models and covers twelve linguistic phenomena such as agreement, binding, and island effects.
EWOK aims to evaluate models' world knowledge and provides a task to match a target text with plausible or implausible contexts.
GLUE provides nine different tasks, which highlight common phenomena such as the use of world knowledge, logical operation, and lexical entailment.

\section{Results and Discussion}
In our experiment, we train GPT-2 from scratch using training data that includes artificial VSs. We report the results after the model training has converged, specifically the results after 3 epochs.

\subsection{Results Using the Sequential Concatenation Method}

\begin{table*}[!ht]
    \begin{adjustbox}{width=\textwidth,center}
    \centering
    \begin{tabular}{ccccccccccc}
    \toprule
        & \multicolumn{2}{c}{BLiMP} & \multicolumn{2}{c}{BLiMP Suppl.} & \multicolumn{2}{c}{EWOK} & \multicolumn{2}{c}{GLUE} & \multicolumn{2}{c}{Macro Avr.} \\ 
        \cmidrule(r){2-3} \cmidrule(lr){4-5} \cmidrule(lr){6-7} \cmidrule(lr){8-9} \cmidrule(l){10-11}
        VS in Dataset & Consec. & Shuf. & Consec. & Shuf. & Consec. & Shuf. & Consec. & Shuf. & Consec. & Shuf. \\
        \midrule \midrule
        0\% & \textbf{60.8} & 61.0 & 56.7 & 57.3 & \textbf{49.9} & \textbf{50.2} & 68.1 & 68.8 & \textbf{58.9} & 59.3 \\
        20\% & 59.0 & 60.6 & 55.8 & 57.5 & 49.1 & 49.5 & 68.7 & \textbf{69.0} & 58.2 & 59.1 \\
        40\% & 58.4 & 60.3 & \textbf{58.3} & 57.6 & 48.4 & 49.7 & 68.8 & 68.2 & 58.5 & 58.9 \\ 
        60\% & 57.9 & 60.9 & 55.6 & \textbf{58.7} & 48.7 & 49.6 & \textbf{69.8} & 68.4 & 58.0 & \textbf{59.4} \\  
        80\% & 57.7 & 60.5 & 56.1 & 57.6 & 48.4 & 49.9 & 69.3 & 67.6 & 57.9 & 58.9 \\ 
        100\% & 57.8 & \textbf{61.7} & 54.8 & 55.4 & 49.3 & 49.6 & 69.6 & 68.8 & 57.9 & 58.9 \\ 
        \bottomrule
    \end{tabular}
    \end{adjustbox}
    \caption{Averaged Scores (\%) of BLiMP, EWOK, and GLUE trained on 3 epochs, where each VS is concatenated into a single sequence. Boldface denotes the highest score per benchmark in each setting. The columns Consec. and Shuf. indicate Consecutive and Shuffle, respectively.}
    \label{tab:results_1}
\end{table*}

Table~\ref{tab:results_1} shows the results for 3 epochs using the dataset containing VSs concatenated into a single line.

First, we focus on the impact of the VS ratio.
In the consecutive condition, the highest macro-average score was achieved when the ratio of VSs was 0\%. Although this contradicts our expectations, it may be due to reduced lexical variation in the training data caused by increased artificial VSs.
Specifically, GLUE scores showed a tendency to improve as the ratio of VSs increased, whereas BLiMP scores declined with the increasing ratio of VSs.
The highest BLiMP Supplement score was achieved with a 40\% ratio of VSs, which aligns closely with the proportion found in real CDS, reflecting a more naturalistic distribution of CDS.
The BLiMP Supplement, like BLiMP, is a binary classification task but focuses on semantic knowledge, including a question-and-answer format. Given the characteristics of this task, VSs are thought to help the model comprehend the meanings of words. VSs consist of a series of sentences with the same meaning but slightly differing in form and structure. Through these patterns, the model can recognize the meaning of each word.
In contrast, the EWOK score was higher at VS ratios of 0\% and 100\%, which differ from the ratio found in actual CDS.

Even under the shuffled condition, scores varied with the VSs proportion. Specifically, the BLiMP score was highest at 100\% VSs, while the BLiMP Supplement score peaked at 60\%. The EWOK score was highest at 0\%, but the difference compared to other proportions was minimal. The GLUE score peaked at 20\% VSs but was nearly the same as at 0\% and 100\%. 
Overall, no consistent trend was observed in the impact of changing the VSs proportion.
Next, we compare the results between the consecutive condition and the shuffle condition.
In all VSs ratio settings, most scores for tasks other than GLUE were higher under the shuffled condition compared to the consecutive condition.
The macro average improved by 0.89\% in the shuffled condition compared to the consecutive condition.
In contrast, for the GLUE scores, the consecutive condition outperformed the shuffled condition when the ratio of VSs was 60\% or higher.
However, a VSs proportion above 50\% diverges from the actual inputs of children, as the highest proportion of VSs in CDS is approximately 50\%.
This discrepancy is because artificial VSs contain less noisy data compared to actual CDS.
CDS contains many fragmentary utterances, as follows:

\begin{enumerate}
    \item[(3)]
    To who? \\
    You don't. \\
    To you or to Laura? \\
    To me. \\
    Oh how come? \\
\end{enumerate}

According to~\citet{Faulkner-etal-2003}, fragments comprise approximately 30\% of CDS. In contrast, artificial CDS contains more full sentences, as follows:

\begin{enumerate}
    \item[(4)]
    It's a blanket that we all share. \\
    We all have a blanket together. \\  
    This blanket belongs to everyone.  \\
    It's a blanket for all of us to use. \\
    Everyone can use this blanket.  
\end{enumerate}

In the BLiMP Supplement, the consecutive condition outperformed the shuffled condition at a 40\% ratio of VSs. 

Overall, these results suggest that using training data where VSs are concatenated into a single line, VSs were effective for GLUE. While BLiMP, BLiMP Supplement, and EWOK are evaluated in a zero-shot setting, GLUE requires fine-tuning. This difference in tasks indicates that the model has not fully acquired grammatical knowledge from VSs alone. However, pre-training using VSs may enhance the efficiency of training for other tasks.
However, contrary to our expectations, it is surprising that the shuffled condition, which disrupts VSs, achieved better scores.

\subsection{Results Using the Adjacent Batch Method}

\begin{table*}[!ht]
    \centering
    \begin{adjustbox}{width=\textwidth,center}
    \begin{tabular}{ccccccccccc}
    \toprule
    & \multicolumn{2}{c}{BLiMP} & \multicolumn{2}{c}{BLiMP Suppl.} & \multicolumn{2}{c}{EWOK} & \multicolumn{2}{c}{GLUE} & \multicolumn{2}{c}{Macro Avr.} \\ 
    \cmidrule(r){2-3} \cmidrule(lr){4-5} \cmidrule(lr){6-7} \cmidrule(lr){8-9} \cmidrule(l){10-11}
    VS in Dataset & Consec. & Shuf. & Consec. & Shuf. & Consec. & Shuf. & Consec. & Shuf. & Consec. & Shuf. \\
        \midrule \midrule
        0\% & 60.8 & 61.0 & 56.7 & 57.3 & \textbf{49.9} & \textbf{50.2} & 68.1 & \textbf{68.8} & 58.9 & 59.3 \\
        20\% & 60.4 & 61.1 & \textbf{59.7} & 58.9 & \textbf{49.9} & 50.1 & 68.4 & 68.1 & \textbf{59.6} & \textbf{59.5} \\ 
        40\% & 60.6 & 60.0 & 58.3 & \textbf{60.2} & 49.3 & 49.5 & \textbf{68.9} & 67.9 & 59.3 & 59.4 \\ 
        60\% & 61.1 & \textbf{61.2} & 58.4 & 58.2 & 49.6 & 49.5 & \textbf{68.9} & 67.6 & 59.5 & 59.1 \\ 
        80\% & 61.5 & 60.8 & 58.8 & 57.8 & 49.6 & 49.5 & 68.6 & 68.0 & \textbf{59.6} & 59.0 \\ 
        100\% & \textbf{61.6} & 61.1 & 57.2 & 57.3 & 49.8 & 49.6 & 68.2 & 67.5 & 59.2 & 58.9 \\
        \bottomrule
    \end{tabular}
    \end{adjustbox}
    \caption{Averaged Scores of BLiMP, EWOK, and GLUE trained on 3 epochs, where each sentence within the VS is placed in adjacent batches. Boldface denotes the highest score per benchmark in each setting. The columns Consec. and Shuf. indicate Consecutive and Shuffle, respectively.}
    \label{tab:results_3}
\end{table*}

Table~\ref{tab:results_3} shows the results of 3 epochs of training using the dataset in which each sentence in the VS is placed in adjacent batches.

First, we focus on the impact of the VS ratio.
Under the consecutive condition, all metrics except for EWOK showed better scores when the VSs were included in the training data. 
Specifically, the BLiMP score increased as the proportion of VSs increased. 
The BLiMP Supplement achieved the highest score when the proportion of VSs was 20\%, which is close to the actual proportion of VSs in CDS.
The GLUE score peaked when the proportion of VSs was 40\% and 60\%, which is slightly higher than the actual proportion in CDS.
Similar to the Sequential Concatenation Method, it is likely that the increase in artificial VSs contributed to reducing noise in the training data.
These results suggest that the optimal proportion of VSs varies depending on the evaluation metric.
While both the BLiMP and GLUE scores benefited from the presence of VSs, the EWOK score was not affected.
Under the shuffled condition, the BLiMP and BLiMP Supplement scores benefited from the presence of VSs in the training data.
The highest scores for each metric were achieved when the VSs proportion was 60\% or lower.
The BLiMP Supplement score increased as the VSs proportion approached the human-like range of 20\%–40\%  under the shuffle condition.

Next, we compare the results between the consecutive condition and the shuffle condition.
For the scores that benefited from the presence of VSs (BLiMP, BLiMP Supplement, GLUE), the scores under the consecutive condition outperformed those under the shuffle condition when the proportion of VSs was optimal for each score.

In summary, with the Adjacent Batch Method, the consecutive condition showed higher scores for metrics other than EWOK when VSs were included in the dataset, indicating the benefit of VSs. However, the shuffled condition still outperformed the consecutive condition in some cases.

\subsection{One-epoch Results}

\begin{table*}[!ht]
    \centering
    \begin{adjustbox}{width=\textwidth,center}
    \begin{tabular}{ccccccccccc}
        \toprule
        & \multicolumn{2}{c}{BLiMP} & \multicolumn{2}{c}{BLiMP Suppl.} & \multicolumn{2}{c}{EWOK} & \multicolumn{2}{c}{GLUE} & \multicolumn{2}{c}{Macro Avr.} \\ 
        \cmidrule(r){2-3} \cmidrule(lr){4-5} \cmidrule(lr){6-7} \cmidrule(lr){8-9} \cmidrule(l){10-11}
        VS in Dataset & Consec. & Shuf. & Consec. & Shuf. & Consec. & Shuf. & Consec. & Shuf. & Consec. & Shuf. \\
        \midrule \midrule
        0\%   & \textbf{58.3} & 58.5 & \textbf{54.7} & 54.9 & \textbf{49.4} & 49.6 & 67.0 & 66.7 & 57.3 & 57.4 \\ 
        20\%  & 57.5 & 58.8 & 54.6 & 53.8 & \textbf{49.4} & 49.5 & 68.7 & 68.3 & 57.5 & 57.6 \\ 
        40\%  & 57.4 & 59.0 & \textbf{54.7} & \textbf{55.2} & 48.8 & 49.2 & 69.7 & 68.8 & \textbf{57.7} & 58.1 \\ 
        60\%  & 57.6 & \textbf{59.4} & 54.1 & 54.9 & \textbf{49.4} & 49.4 & 69.2 & \textbf{70.0} & 57.6 & \textbf{58.4} \\ 
        80\%  & 57.0 & 58.8 & 54.6 & 54.6 & 49.0 & 49.4 & 69.7 & 69.1 & 57.6 & 58.0 \\
        100\% & 56.6 & 58.7 & 53.3 & 54.9 & 49.0 & \textbf{49.8} & \textbf{70.2} & 69.2 & 57.3 & 58.1 \\ 
        \bottomrule
    \end{tabular}
    \end{adjustbox}
    \caption{Averaged Scores (\%) of BLiMP, EWOK, and GLUE trained on 1 epoch, where each VS is concatenated into a single sequence. Boldface denotes the highest score per benchmark in each setting. The columns Consec. and Shuf. indicate Consecutive and Shuffle, respectively.}
    \label{tab:results_2}
\end{table*}

\begin{table*}[!ht]
    \centering
    \begin{adjustbox}{width=\textwidth,center}
    \begin{tabular}{ccccccccccc}
        \toprule
        & \multicolumn{2}{c}{BLiMP} & \multicolumn{2}{c}{BLiMP Suppl.} & \multicolumn{2}{c}{EWOK} & \multicolumn{2}{c}{GLUE} & \multicolumn{2}{c}{Macro Avr.} \\ 
        \cmidrule(r){2-3} \cmidrule(lr){4-5} \cmidrule(lr){6-7} \cmidrule(lr){8-9} \cmidrule(l){10-11}
        VS in Dataset & Consec. & Shuf. & Consec. & Shuf. & Consec. & Shuf. & Consec. & Shuf. & Consec. & Shuf. \\
        \midrule \midrule
        0\%   & 58.3 & 58.5 & 54.7 & 54.9 & 49.4 & 49.6 & 67.0 & 66.7 & 57.3 & 57.4 \\
        20\%  & 58.8 & 59.0 & 55.6 & 55.3 & \textbf{49.5} & 49.6 & 68.3 & 68.7 & 58.1 & 58.1 \\
        40\%  & 59.1 & 59.1 & 53.8 & 54.8 & \textbf{49.5} & 49.3 & 68.4 & 67.5 & 57.7 & 57.7 \\
        60\%  & \textbf{59.5} & \textbf{59.2} & 54.1 & \textbf{55.9} & 49.1 & 49.2 & 68.6 & \textbf{70.6} & 57.8 & \textbf{58.7} \\
        80\%  & 59.3 & 59.1 & 55.5 & 55.5 & \textbf{49.5} & 49.4 & \textbf{69.2} & 69.2 & \textbf{58.4} & 58.3 \\
        100\% & 58.6 & 58.5 & \textbf{55.9} & 55.4 & 49.2 & \textbf{49.8} & 68.1 & 68.3 & 57.9 & 58.0 \\
        \bottomrule
    \end{tabular}
    \end{adjustbox}
    \caption{Averaged Scores of BLiMP, EWOK, and GLUE trained on 1 epoch, where each sentence within the VS is placed in adjacent batches. Boldface denotes the highest score per benchmark in each setting. The columns Consec. and Shuf. indicate Consecutive and Shuffle, respectively.}
    \label{tab:results_4}
\end{table*}

While models observe the same instances multiple times by training on multiple epochs, children only see a single instance only one time in natural speech interaction. 
VS has a role in increasing the salience of structural properties that are hard to recognize from a single instance, thereby exposing children to instances that have the same semantic intentions with slightly different words and structures.
Therefore, there is a possibility that the effect of VSs diminishes when training over multiple epochs.
We report results after training for only 1 epoch to examine this possibility.

The results for one epoch training using the Sequential Concatenation and Adjacent Batch Method are shown in Tables~\ref{tab:results_2} and~\ref{tab:results_4}.
Regarding the impact of the VSs proportion in the training data, with the Sequential Concatenation Method, the impact of VSs proportion was similar to that in the 3-epoch training: BLiMP scores decreased as the VSs proportion increased, while GLUE scores improved. The highest scores for BLiMP, BLiMP Supplement, and GLUE were observed within the 40\%–60\% range.
With the Adjacent Batch Method, the highest scores for each metric were achieved when the VSs proportion was 60\% or higher. 

Regarding the differences in results between the shuffled and consecutive conditions, in the Sequential Concatenation Method results, the GLUE score was higher in the consecutive condition than the shuffled condition, except at 60\% VSs. However, for most other metrics, the shuffled condition outperformed the consecutive condition.
Similarly, in the Adjacent Batch Method results, none of the metrics showed a significant advantage for the consecutive condition over the shuffled condition.

Contrary to our expectations, the effects of VSs were not more pronounced in the 1-epoch training compared to the 3-epoch training.

\subsection{Discussion}
Taken together, our results show that the presence of CDS-inspired variations is often beneficial. However, ---somewhat counterintuitively--- presenting this variation in a shuffled order is often better than presenting them consecutively as in CDS.
An additional finding is that the optimal amount of VSs varies among settings and evaluation benchmarks, and we could not find an overall winner.
This might be due to the fact that, in our current experimental design, the amount of VSs is in direct competition with the diversity of utterances present in the dataset (i.e. potentially larger coverage of vocabulary and constructions in the datasets with less VSs).
To better disentangle the effect of variations from that of corpus diversity, we are currently planning an experiment where a given amount of variations will be compared to a similar amount of identical repetitions.

\section{Conclusion}
We presented an initial exploration of the effect of CDS-inspired variation sets on language model training efficiency. 
Our results suggest that VSs can have a beneficial impact on various linguistic competences. 
They also reveal that this effect is entrenched with several factors like the order of utterance exposure and the number of training epochs, leaving space for more detailed investigations in the future.

\section{Limitations}
There are several limitations in this research.
\texttt{gpt4o-mini} does not necessarily generate VSs that closely resemble natural VSs.
Consequently, there is a possibility that our training data may contain unintended noise.
Furthermore, we shuffled CDS to fully control the number of VSs in the training dataset. This procedure disrupted the natural VSs in CDS, possibly affecting the scores negatively.
Additionally, the vocabulary size could not be strictly aligned between the ``Sequential Concatenation Method'' and the ``Adjacent Batch Method.'' While the difference in vocabulary size is marginal, it may influence the scores.

\section{Ethics Statement}
This study was conducted in accordance with ethical guidelines and regulations.
We utilized natural speech data extracted from CHILDES~\citep{macwhinney2000}. This is an open source corpus that archives natural speech between caregivers and their children. The data are archived without confidential information about the participants as children are usually given pseudonyms. 
Following the ACL Policy on Publication Ethics, we used ChatGPT to assist in refining the wording. We also partially relied on ChatGPT to generate code for prepossessing and evaluation.

\section*{Acknowledgments}
This work was supported by JSPS KAKENHI Grant Number 24H00087, JST PRESTO Grant Number JPMJPR21C2 and ACT-X Grant Number JPMJAX24CM.
Arianna Bisazza was supported by the Dutch Research Council (NWO) within the Talent Programme (VI.Vidi.221C.009). 

\bibliography{custom}

\appendix

\section{Prompt for Generating Artificial VSs}
\label{sec:appendix-prompt}
To generate synthesis VSs, we used the following prompt:

\begin{quote}
Rephrase a given sentence based on the characteristics of variation sets. A variation set is a set of utterances that have characteristics as follows: \\
In successive utterances, \\
- the same content is repeated or rephrased. \\
- there is a consistent intent. \\
- there are operations such as word substitution, addition/deletion of phrases, and reordering of phrases. \\
Here is an example: \\
You can put the animals there.\\
You can take the pig and the cat and put them there.\\
Can you put them there?\\
Good.\\
Can you put the pig there too?\\
Please use only the vocabulary that 10 year-old children understand.
\end{quote}

\section{Hyperparameters} \label{tab:hypara}
\begin{table}[htbp]
    \centering
    \begin{tabular}{@{}llc@{}}
        \toprule
        \multirow{8}{*}{Model}  
        & architecture & GPT-2 \\
        & parameters & 124M \\
        & vocab size & 50,257 \\
        & hidden size & 768 \\
        & heads & 12 \\
        & layers & 12 \\
        & dropout & 0.1 \\
        & layer norm eps & 1e-05 \\
        & initializer range & 0.02 \\
        \cmidrule(r){1-1} \cmidrule(lr){2-2} \cmidrule(lr){3-3}
        \multirow{4}{*}{Optimizer}               
        & algorithm & AdamW \\
        & learning rates & 5e-05 \\
        & betas & (0.9, 0.999) \\
        & weight decay & 0.0 \\
        \cmidrule(r){1-1} \cmidrule(lr){2-2} \cmidrule(lr){3-3}
        Scheduler
        & type  & linear \\
        \cmidrule(r){1-1} \cmidrule(lr){2-2} \cmidrule(lr){3-3}
        \multirow{5}{*}{Training}                
        & gradient accumulation & 1 \\
        & epoch & 3 \\
        & batch size & 64 \\
        & line by line & true \\
        & NGPU & 1 \\
        \bottomrule
    \end{tabular}
\caption{Hyperparameters of the language models.}
\end{table}

\end{document}